\documentclass[a4paper]{article}

\usepackage{INTERSPEECH2020,multirow}

\title{Cross-Utterance Language Models with Acoustic Error Sampling}
\name{G. Sun, C. Zhang and P. C. Woodland \thanks{Thanks to Yu Wang for his help with the Kaldi acoustic model. G. Sun is funded by Cambridge Trust.}}
\address{Cambridge University Engineering Dept., Trumpington St., Cambridge, CB2 1PZ U.K}
\email{\{gs534, cz277, pcw\}@cam.ac.uk}

\begin{document}

\maketitle
\begin{abstract}

\end{abstract}

The effective exploitation of richer contextual information in language models (LMs) is a long-standing research problem for automatic speech recognition (ASR). A cross-utterance LM (CULM) is proposed in this paper, which augments the input to a standard long short-term memory (LSTM) LM with a context vector derived from past and future utterances using an extraction network. 
%A cross-utterance neural network LM is proposed in this paper, which consists of a standard long short-term memory (LSTM) LM with an augmented context vector input and a jointly trained context vector extraction network. The context vector is generated from the past and future utterances surrounding the current one using the extraction network. 
%which consists of an extraction network and a standard LM, whose  
%uses a context vector as an auxiliary input feature to adapt the LM. The context vector is extracted from the past and future utterances surrounding the current one through an extraction network. 
%Two types of extraction networks are proposed. The first one uses another LSTM to encode each surrounding utterance into a memory cell vector, and transform the concatenation of the vectors with a projection. The second one includes an additional multi-head self-attentive layer on top of the LSTM encoding. 
The extraction network uses another LSTM to encode surrounding utterances into vectors which are integrated into a context vector using either a projection of LSTM final hidden states, or a multi-head self-attentive layer. In addition, an acoustic error sampling technique is proposed to reduce the mismatch between training and test-time. This is achieved by considering possible ASR errors into the model training procedure, and can therefore improve the word error rate (WER).
Experiments performed on both AMI and Switchboard datasets show that CULMs outperform the LSTM LM baseline WER.
In particular, the CULM with a self-attentive layer-based extraction network and acoustic error sampling achieves 0.6\% absolute WER reduction on AMI, 0.3\% WER reduction on the Switchboard part and 0.9\% WER reduction on the Callhome part of \emph{Eval2000} test set over the respective baselines.


\section{Introduction}

A language model (LM) estimates the probability of a word sequence which is often decomposed into a product of conditional word prediction probabilities using the chain rule. LMs are widely used in many machine learning tasks, such as natural language understanding, machine translation and automatic speech recognition (ASR). In ASR, high-performance LMs are found to be critical to achieve good performance
for both traditional source-channel model-based systems \cite{Cmap, LMinASR2} and recent end-to-end systems \cite{LMinASR3, LMinASR4, LMinASR5}. While traditional $n$-gram LMs can only provide the contextual information from $n$ preceding words \cite{Cngram}, recurrent neural network (RNN) LMs \cite{LMinASR7,AndrewLiuRNNLM}, particularly long short-term memory (LSTM) LMs \cite{LSTM,C1}, can provide richer contextual information from the entire utterance to achieve better ASR performance \cite{LSTMLM1, LSTMLM2}. More recently, contextual information from past and future utterances has been taken into account in language modelling \cite{TransformerXL,MShumanparity,BrianThesis}.

%LMs provide the prior information for the \textit{maximum a posteriori} (MAP) decoding \cite{Cmap}. Recently, long short-term memory (LSTM) LMs \cite{Cngram, C1, C2} have shown significant improvements over traditional n-gram LMs. When incorporated into ASR systems in the second-pass rescoring process, these LMs are usually trained and used at single utterance level. Nevertheless, the context information may be greatly beneficial to the modelling task in applications such as meeting and conversation transcription systems \cite{C13, C14, C15}.

Although effective incorporation of cross-utterance contextual information remains an open research problem, improvements have been found in ASR performance using such information.
%While the effective incorporation of such cross-utterance contextual information still remains an open research problem, improvements are found in ASR performance using contextual information.
Early work tried to incorporate a short-term cache \cite{C4} or document-level semantic information \cite{C5} into LMs. With the advent of RNN LMs, model adaptation using statistical analysis to represent global topics \cite{C3,C7} also showed promising results. %More recently, neural networks are used to encode contextual information into vector representations for LM adaptation \cite{C6,C9,C10,Cdeepcont}
More recently, neural networks, such as hierarchical RNNs and pre-trained BERT LMs \cite{C11}, were used to encode contextual information into vector representations for LM adaptation \cite{C6,C9,C10,Cdeepcont}.
%\cite{C9} explores the use of hierarchical RNN structure, and \cite{C10} adopts the context encoding from a pre-trained bidirectional transformer (BERT) \cite{C11} model. 
To use the history information from the previous utterances at test-time, an RNN LM was trained with conversational data without resetting its recurrent state at the beginning of each new utterance \cite{C8}.
%RNN recurrent vectors for history representations during decoding, \cite{C8} uses RNNLMs at conversation level without re-initialising the RNN hidden state at the beginning of each utterance.
%On the other hand, to take the advantage of the RNN state for history representations during decoding, \cite{C8} uses RNNLMs at conversation level without re-initialising the RNN hidden state at the beginning of each utterance. 
%Moreover, an attention-based history representation concatenated with the hidden states is proposed for LMs trained on long context \cite{C12, C22}.
Moreover, the history representation can be extracted with an attention model and concatenated with the hidden states \cite{C12, C22}.

This paper proposes an LSTM-based cross-utterance LM (CULM) structure to explore the flexible and effective use of context information embedded in the past and future utterances. The CULM comprises of a main LSTM and a context extraction network \cite{C7}. The main LSTM takes a context vector as an auxiliary input on top of a standard LSTM LM structure. The context vector encapsulating information from surrounding utterances is generated from the extraction network which contains an encoder LSTM and a fusion component.
To extract the context vector, the surrounding utterances are first split into short segments and encoded into vectors using the encoder LSTM. The fusion component is then used to transform these vectors into the context vector. Two extraction methods, the final-state encoding and the self-attentive encoding, have been investigated. In \textit{final-state encoding}, a fully-connected (FC) layer is used to transform the concatenation of the encoder LSTM final output states corresponding to each segment into the context vector. Alternatively, the \textit{self-attentive encoding} employs a self-attentive layer \cite{C16} to fuse the entire encoder LSTM output sequence instead of only the final state for each segment before sending it to the FC layer. The main LSTM and the context extractor are jointly optimised so that the extractor learns useful information for predicting the next word.
%two context extraction networks are investigated in this paper. 
%The first one takes the final output state of the LSTM on each segment, and extracts the context vector using a fully-connected layer on the concatenation of these output states. This is referred to as the final-state encoding. Alternatively, the second one applies a multi-head self-attentive layer \cite{C16} to the sequence of LSTM outputs to encode the context, which is referred to self-attentive encoding. The LSTM-LM and the context extraction network are jointly optimised so that the extraction network learns to incorporate useful information for next-word prediction. 

Since in real ASR applications, only decoding hypotheses of the surrounding utterances are available at test-time,
the CULM trained on the reference text of surrounding utterances has a mismatch between training and test \cite{C20, C21}. Therefore, an acoustic error sampling method is proposed to introduce simulated ASR errors into surrounding utterances in training. Experiments were performed on both augmented multi-party interaction (AMI) and Switchboard corpora, and results are measured in both (pseudo) perplexity (PPL) and word error rate (WER).

The remainder of this paper is as follows: Sec. \ref{sec:2} describes the CULM structure, followed by the introduction of the two context extraction networks. Acoustic error sampling method is explained in Sec. \ref{sec:4}. Experiments and results are discussed in Sec. \ref{sec:5} followed by conclusions in Sec. \ref{sec:6}.

\section{CULM structure}
\label{sec:2}

\begin{figure}[h]
    \centering
    \includegraphics[scale=0.38]{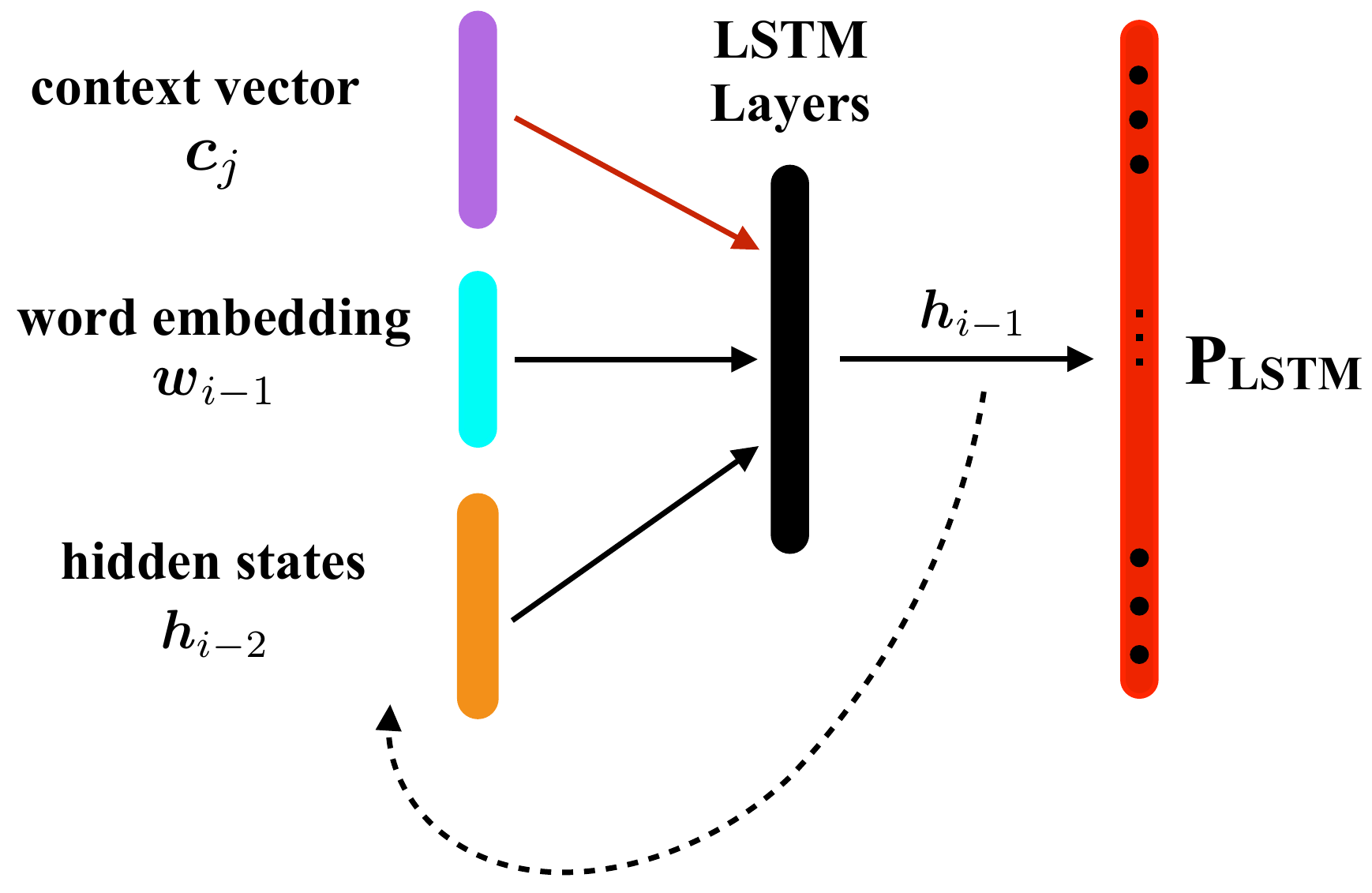}
    \vspace{-0.2cm}
    \caption{CULM structure predicting the i-th word in the j-th utterance. Hidden state $\mathbf{h}_{i-1}$ is carried to the next prediction. $\text{P}_{\text{LSTM}}$ is the model output probability distribution for the next word. $\mathbf{w}$ and $\mathbf{c}$ are used to represent word embeddings and context vectors, while $w$ and $c$ are used for words and contexts.}
    \label{fig:model}
    \vspace{-0.5cm}
\end{figure}

The model structure is shown in Fig. \ref{fig:model} which includes the context vector as an auxiliary input to the main LSTM \cite{C18}. When processing the $j$-th utterance, a context vector is extracted from the surrounding utterances excluding the current one. The input to the main LSTM is the concatenation of the context vector, the word embedding and hidden states. The output is a probability distribution over the vocabulary for the i-th word given previous words in the utterance and the context of the utterance, $\text{P}_{\text{LSTM}} = \text{P}({w}^j_i|{w}^j_{1:i-1}, {c}_j)$. LSTM hidden states are re-initialised at the beginning of each utterance to avoid information leaking from the content of previous utterances to the current utterance when future information is used. The model together with the context vector extraction network is jointly trained using the cross-entropy criterion.

\subsection{Extracting information from surrounding utterances}
\label{sec:3}

\subsubsection{Final-state encoding}
\label{sec:lstmenc}
\begin{figure}[h]
    \centering
    \includegraphics[scale=0.52]{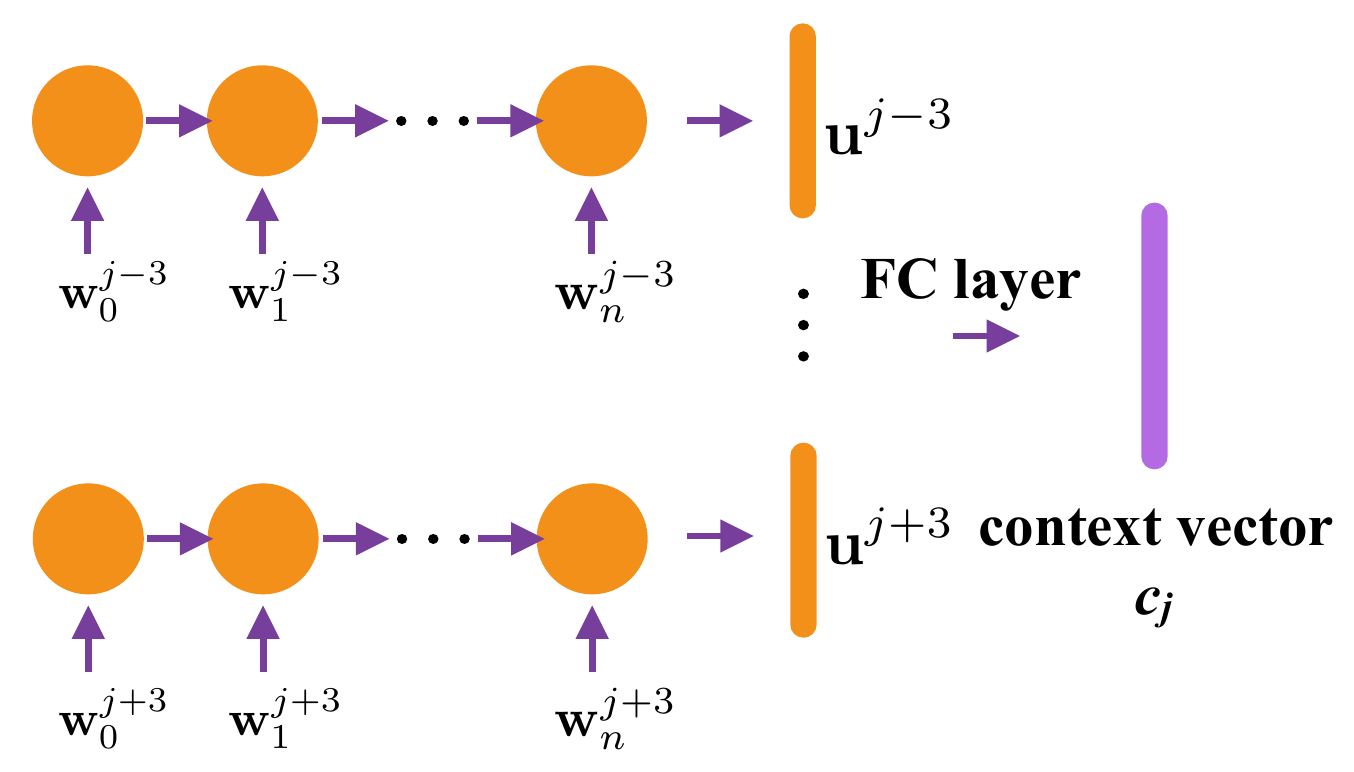}
    \caption{Context vector extraction using the encoder LSTM final hidden state. Three past and future segments relative to the current utterance $j$ are shown here for illustration. FC is the fully-connected layer and $\mathbf{u}$'s are LSTM final state encodings.}
    \label{fig:extract1}
    \vspace{-0.3cm}
\end{figure}

 The final-state encoding extraction method is shown in Fig. \ref{fig:extract1} where the segments from the context are directly encoded by the final hidden state of the encoder LSTM.

 When training the LSTM, the output gate learns to fetch useful information for specific tasks from the \textit{cell} state which is the memory in the network. The final-state encoding method takes advantage of this design and encourages the encoder LSTM final hidden state to extract useful context information. First, the context is arranged by concatenating the past and future utterances separately, excluding the current one. In contrast to \cite{C23}, where the utterance structure is retained, segments are obtained by splitting the context into sequences with equal number of words ignoring utterance boundaries. This facilitates parallel processing when the extraction network is jointly optimised with the LM network. Then, each segment is processed by an LSTM and the final hidden state, $\mathbf{u}$, is used as the segment encoding. Finally, segment encodings are concatenated and projected through a fully-connected layer to form the context vector $\mathbf{c}_j$ in Eqn (\ref{eq:fc}).
\vspace{-0.0cm}
\begin{equation}
    \mathbf{c}_j = \text{ReLU}(\mathbf{W}^T[\mathbf{u}_{j-l}, ..., \mathbf{u}_{j-1}, \mathbf{u}_{j+1}, ..., \mathbf{u}_{j+k}]).
    \label{eq:fc}
\end{equation}

\subsubsection{Self-attentive encoding}
\label{sec:selfatten}

\begin{figure}[h]
    \centering
    \includegraphics[scale=0.55]{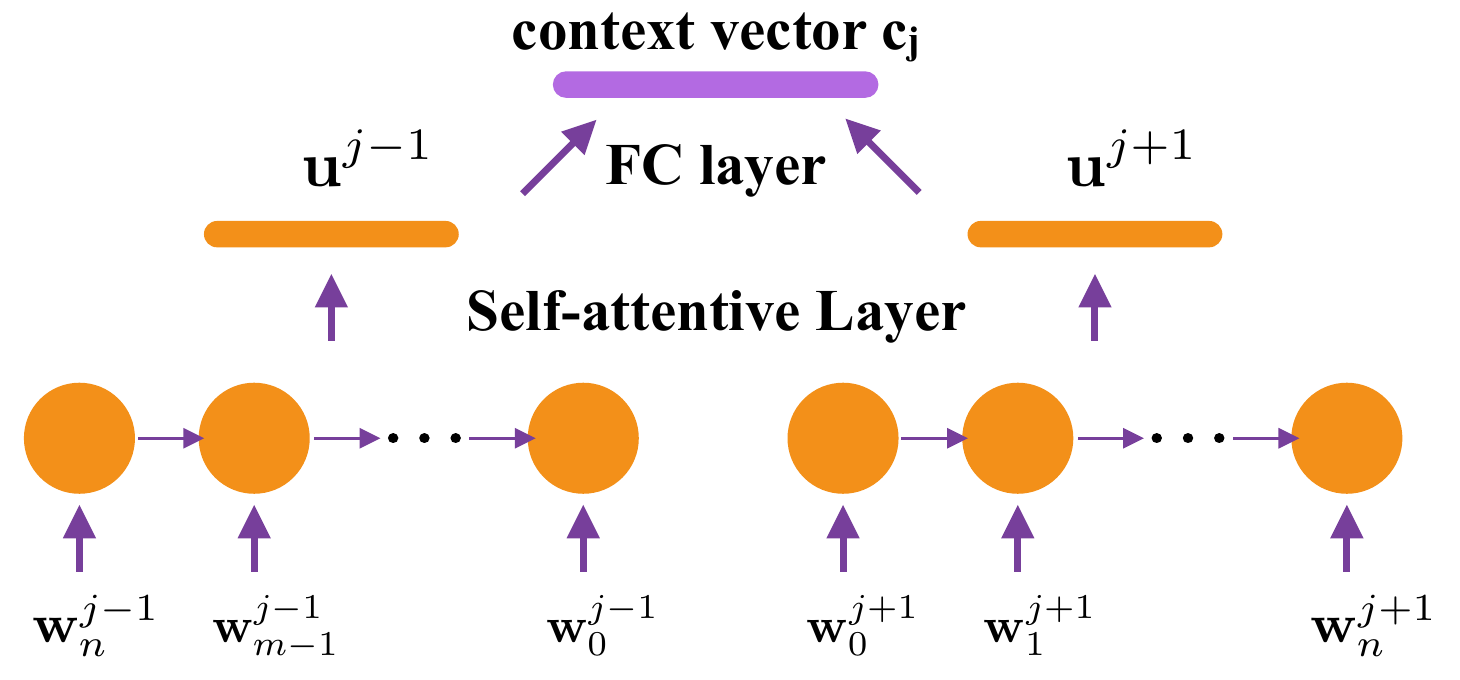}
    \caption{Context vector extraction using multi-head self-attentive segment encodings. One past segment and one future segment, each with n words, are shown.}
    \label{fig:extract2}
    \vspace{-0.3cm}
\end{figure}

The final hidden state has a limited ability to capture the content of the entire segment and may easily have a bias towards the ending words.
Therefore, a self-attentive layer is introduced to enrich the representation and explicitly indicate the relative importance among context words, as illustrated in Fig. \ref{fig:extract2}.

 As before, the context is arranged into segments which are sent to the encoder LSTM. The entire sequence of hidden states from the encoder LSTM is then combined by a multi-head self-attentive layer \cite{C16, C17}. The self-attentive layer combines the sequence of output hidden states using a weighted average where the attention combination weights are generated through a two-layer feed-forward neural network from the hidden state sequence itself. The relative importance of each word in the context is directly reflected by the attention combination weight it is assigned. This network extracts the past and future segment encodings separately, and fuses them together with a fully-connected layer whose output is the context vector. This process is shown in Eqn (\ref{eq:atten}).
\begin{align}
    \mathbf{u}^{j-1} &= \text{SelfAtten}[\text{LSTM}(\mathbf{w}_{n}^{j-1}, ..., \mathbf{w}_{0}^{j-1})], \nonumber\\
    \mathbf{u}^{j+1} &= \text{SelfAtten}[\text{LSTM}(\mathbf{w}_{0}^{j+1}, ..., \mathbf{w}_{n}^{j+1})], \nonumber\\
    \mathbf{c}_j &= \text{FC}(\mathbf{u}^{j-1}, \mathbf{u}^{j+1}),
    \label{eq:atten}
\end{align}
where $\mathbf{u}$'s represent the self-attentive segment encodings. $\text{LSTM}(\cdot)$ refers to the encoder LSTM. $\text{SelfAtten}(\cdot)$ refers to the self-attentive layer in \cite{C16} with a multi-head output to generate the context vector $\mathbf{c}_j$. $\text{FC}(\cdot)$ refers to the fully-connected layer.

\section{Acoustic error sampling for context}
\label{sec:4}

% \begin{equation}
%     \hat{\mathbf{c}}_j \sim \text{P}(\mathbf{c}_j|\mathbf{O}_{j})
% \end{equation}

In order to reduce the difference between LM training and decoding, insertion, deletion and substitution errors are added to the context before the extraction, which is referred to as acoustic error sampling. This process occurs at the beginning of each training epoch. Each word in the training set will be either deleted, substituted, or inserted with a following word according to an error distribution as shown in Eqn. (\ref{eq:samp}).
\vspace{-0.2cm}
\begin{equation}
    \hat{{c}}_j \sim P({c}_j|\mathbf{O}),
    \label{eq:samp}
    \vspace{-0.2cm}
\end{equation}
where ${c}_j$ is the context of utterance $j$ and $\mathbf{O}$ is the acoustic observation of the context. Two ways of approximating the error distribution are proposed. The first one manually sets the probability of occurrence for each error type, and adopts a uniform distribution across all words for substitutions and insertions. The second one samples according to the error analysis of the training set after first-pass decoding. For instance, in the first method, each word will be deleted, substituted, or inserted with any other word, with probabilities 0.10, 0.08 and 0.04. In the second method, if the word "they" in the training set has 1000 occurrences where 30 of them are deleted and 50 are substituted with "he", the sampling distribution will be 0.03 for deletion and 0.05 for substitution with "he". As before, it also has a probability of 0.04 to be inserted with a word according to the table of insertion frequency counts.

\section{Experiments}
\label{sec:5}

\subsection{Experimental setup}
\subsubsection{Data}

The proposed techniques were evaluated by performing experiments on two tasks: AMI \cite{C19} and the Switchboard (SWBD). AMI contains 100 hours of group meetings recorded at different sites while SWBD contains conversations between two people on defined topics. Data size and partition used in LM training is shown in Table. \ref{tab:data}. The LM train set used for the SWBD task combines data from the SWBD and Fisher corpora which also contains conversations between two people. 10\% of conversations were randomly selected to form the held-out validation set. \textit{Eval2000} test set, which contains conversations from Callhome (CH) and SWBD, is used to evaluate the SWBD system.

\begin{table}[h]
    \centering
    \begin{tabular}{ccccc}
         \toprule
         \textbf{Data} & \textbf{Train} & \textbf{Validation} & \textbf{Test} & \textbf{Vocabulary} \\
         \midrule
         AMI & 911K & 108K & 103K & 13K \\
         SWBD & 24M & 3M & 0.05M & 30K \\
         \bottomrule
    \end{tabular}
    \vspace{0.2cm}
    \caption{Number of words in each split and the vocabulary.}
    \label{tab:data}
    \vspace{-0.2cm}
\end{table}
\vspace{-0.7cm}
\subsubsection{Evaluation}
The performance of the proposed CULMs are measured by pseudo-PPL and WER. The pseudo-PPL is defined in Eqn. \ref{eq:ppl}.
\vspace{-0.2cm}
\begin{equation}
    \text{log}_2(\text{pseudo-PPL}) = {-\frac{1}{N}\sum_{j=1}^{M}\sum_{i=1}^{N_j} \text{log P}({w}^j_i|{w}^j_{1:i-1}, {c}_j)},
    \label{eq:ppl}
    \vspace{-0.2cm}
\end{equation}
where $N=MN_j$ is the total number of words in the test set which contains $M$ utterances and the j-th utterance has $N_j$ words. $-\text{log P}({w}^j_i|{w}^j_{1:i-1}, {c}_j)$ is the cross-entropy loss for each word prediction. It is different to standard PPL in that the context contains future words. However, when hidden state re-initialisation is applied at utterance boundaries and the context excludes the current utterance, the pseudo-PPL reflects the model performance of individual utterances in the test set.

\subsection{AMI Experiments}
The first set of experiments were performed on AMI. Single-layer LSTMs implemented in PyTorch \cite{Cpytorch} with 768d hidden states and 256d word embeddings were used, and were trained on in-domain data only, using stochastic gradient descent (SGD) algorithm with a newbob training scheduler. For WER experiments, a TDNN-F \cite{Ctdnnf} acoustic model was trained on the 81 hours of AMI training data using the lattice-free maximum mutual information \cite{Clfmmi} criterion without any data-augmentation, speaker adaptation or voice tract length normalisation following the simplified Kaldi recipe \cite{Ckaldi,AMAMI}. It was used to generate 100-best lists with a 4-gram LM. The 100-best list were then rescored using CULMs whose context is derived from the hypothesis rescored by a standard LSTM-LM. The code for LM training and rescoring can be found here\footnote{ \texttt{https://github.com/BriansIDP/Cross\_Utterance\_Clean}}.

\subsubsection{Context coverage investigation}

\begin{figure}[t]
    \centering
    \hspace{-0.3cm}
    \includegraphics[scale=0.57]{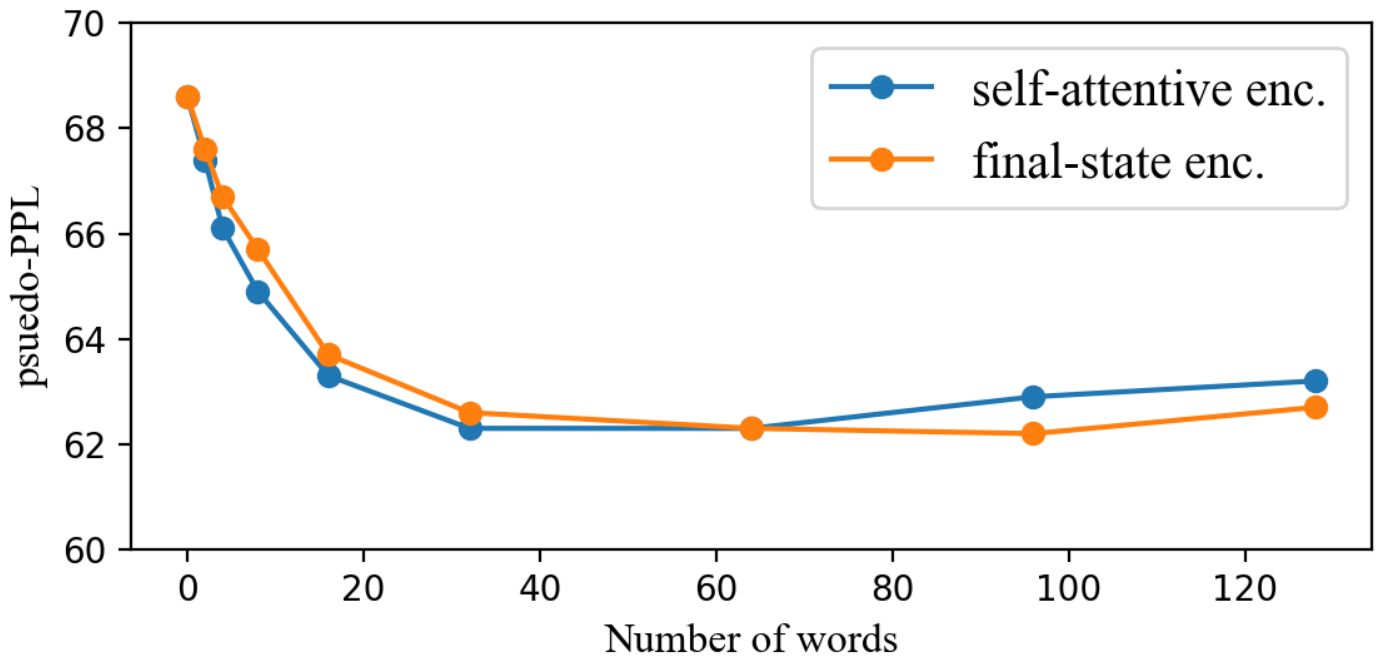}
    \caption{pseudo-PPL by varying the context range on AMI dev set. The context range is measured by the number of words in both past and future contexts.}
    \vspace{-0.5cm}
    \label{fig:PPLcontext}
\end{figure}
To determine suitable context coverages for CULMs, pseudo-PPLs are plotted against different context ranges for the AMI dev set in Figure \ref{fig:PPLcontext}. The lowest pseudo-PPL achieved by the self-attentive encoding is at around 36 words, while the lowest pseudo-PPL for the final-state encoding is at around 72 words. Furthermore, the self-attentive context encoding degrades faster than the final-state encoding when the context becomes longer, because the self-attentive structure is also possible to assign high weights to further context irrelevant to the current utterance. As a result, 36 words for self-attentive encoding and 72 words for final-state encoding were selected.

\subsubsection{System comparison}

Using the best context range, the LM performance for both PPL and WER are compared in Table \ref{tab:sys1}. Baseline refers to the standard single layer LSTM-LM with 768d hidden states and 256d word embeddings. For completeness, using LSTM hidden states from 1-best history to initialise the baseline model at the start of each utterance during rescore is also included which is a simple but powerful incorporation of context information. The corresponding PPL for this is to use the final hidden state from the previous utterance at the beginning of the current utterance at the sentence boundaries without resetting.

\begin{table}[h]
    \centering
    \begin{tabular}{ccclll}
    \toprule
         \textbf{Systems} & \textbf{(pseudo-) PPL} & \textbf{WER} \\
         \midrule
         4-gram & - & 20.2 \\
         baseline & 61.3 & 18.4 \\
         baseline + 1-best & 56.9 & 18.0 \\
         \midrule
         final-state encoding & 55.9 & 18.0\\
         self-attentive encoding & \textbf{55.8} & \textbf{17.9}\\
         \bottomrule
    \end{tabular}
    \vspace{0.2cm}
    \caption{Comparison among different LMs for (pseudo-)PPL and WER on AMI eval set. Baseline + 1-best uses 1-best history to initialise the LSTM hidden states during rescoring.}
    \vspace{-0.5cm}
    \label{tab:sys1}
\end{table}
% \vspace{-0.5cm}
By initialising with 1-best history, PPL is reduced by 4.4 and WER is reduced by 0.4\% absolute compared to the baseline. Both CULMs outperform the baseline the baseline with 1-best history. While requiring more computation than the final-state encoding, the self-attentive encoding CULM achieves the best performance in pseudo-PPL and WER with a further 0.1\% WER reduction over the baseline with 1-best history.

The improvements from using acoustic error sampling technique are shown in Table \ref{tab:sampppl} and Table \ref{tab:sampwer} for PPL and WER respectively. The middle rows in both tables are the results using the first error sampling method and the bottom rows are using the second method with error distributions from analysis files.
\begin{table}[t]
    \centering
    \begin{tabular}{ccclll}
    \toprule
         \textbf{Sampling method} & \textbf{Final-state} & \textbf{Self-atten.} \\
         \midrule
         no sampling & 55.9 & 55.8\\
         \midrule
         substitution & 56.1 & 55.9 \\
         + insertion \& deletion & 55.9 & 55.4\\
         \midrule
         from error analysis & \textbf{55.5} & \textbf{55.3}\\
         \bottomrule
    \end{tabular}
    \vspace{0.2cm}
    \caption{pseudo-PPL with error sampling under two context vector extraction networks on AMI eval set. Pseudo-PPLs are measured under the true context.}
    \label{tab:sampppl}
    \vspace{-0.3cm}
\end{table}
\begin{table}[t]
    \centering
    \begin{tabular}{ccclll}
    \toprule
         \textbf{Sampling method} & \textbf{Final-state} & \textbf{Self-atten.} \\
         \midrule
         no sampling & 18.0 & 17.9 \\
         no sampling true context & 17.9 & 17.9\\
         \midrule
         substitution & 18.0 & 18.0 \\
          + insertion \& deletion & 17.9 & 17.8\\
         \midrule
         from error analysis & \textbf{17.9} & \textbf{17.8}\\
         \bottomrule
    \end{tabular}
    \vspace{0.2cm}
    \caption{WER with error sampling using two context vector extraction networks on AMI eval set.}
    \vspace{-0.8cm}
    \label{tab:sampwer}
\end{table}
% \vspace{-0.3cm}
  Adding insertion and deletion errors that change the position of words in the context performs better than using substitution errors only. By using the second method, both systems achieved another 0.1\% absolute WER reduction. In particular, the self-attentive context encoding system performs better than the no sampling case even when the true context is given. This means the model with error sampling generalises better to other contexts. Therefore, the error sampling not only narrows the gap between training and rescoring, but also has a data augmentation effect to the self-attentive context extraction network.

\subsection{SWBD Experiments}
To further validate the proposed method on a larger corpus, experiments are performed on Switchboard. A two-layer LSTM with 256d word embedding and 2048d hidden states is used as the baseline LM and also for the CULM. Around 260 hours of SWBD training data was used to train the acoustic model with other set-up the same as for AMI \cite{AMSWBD}. 

\subsubsection{System comparison}
Different LMs are compared in Table \ref{tab:sys2} where similar improvements to the AMI case were found. Besides, the effect of error sampling is more significant for the CH part than for the SWBD part as the latter has context closer to the ground truth. Overall, reductions of 0.3\% and 0.9\% absolute WER were found by using a CULM with self-attentive encoding and acoustic error sampling on the SWBD and CH parts respectively.
\begin{table}[t]
    \centering
    \begin{tabular}{ccccllll}
    \toprule
         \multirow{2}{*}{\textbf{Systems}} & \textbf{(pseudo-)} & \multicolumn{2}{c}{\textbf{WER}} \\
         & \textbf{PPL}& \textbf{SWBD} & \textbf{CH} \\
         \midrule
         4-gram & 87.1 & 9.9 & 20.7 \\
         baseline & 52.1 & 7.8 & 17.8\\
         baseline + 1-best & 39.9 & 7.7 & 17.1\\
         \midrule
         final-state enc. & 40.1 & 7.6 & 17.1 \\
         self-atten. & 37.9 & \textbf{7.5} & 17.0 \\
         \midrule 
         final-state enc. + sampling & 39.8 & 7.6 & 17.1 \\
         self-atten. + sampling & \textbf{37.7} & \textbf{7.5} & \textbf{16.9}\\
         \bottomrule
    \end{tabular}
    \vspace{0.2cm}
    \caption{PPL and WER for LM trained on SWBD + Fisher datasets and evaluated on the Eval2000 dataset which is split into the Switchboard (SWBD) part and the Callhome (CH) part. Pseudo-PPLs are measured with the true context.}
    \vspace{-0.8cm}
    \label{tab:sys2}
\end{table}

\subsubsection{Error analysis}
\begin{figure}[h]
    \centering
    \includegraphics[scale=0.3]{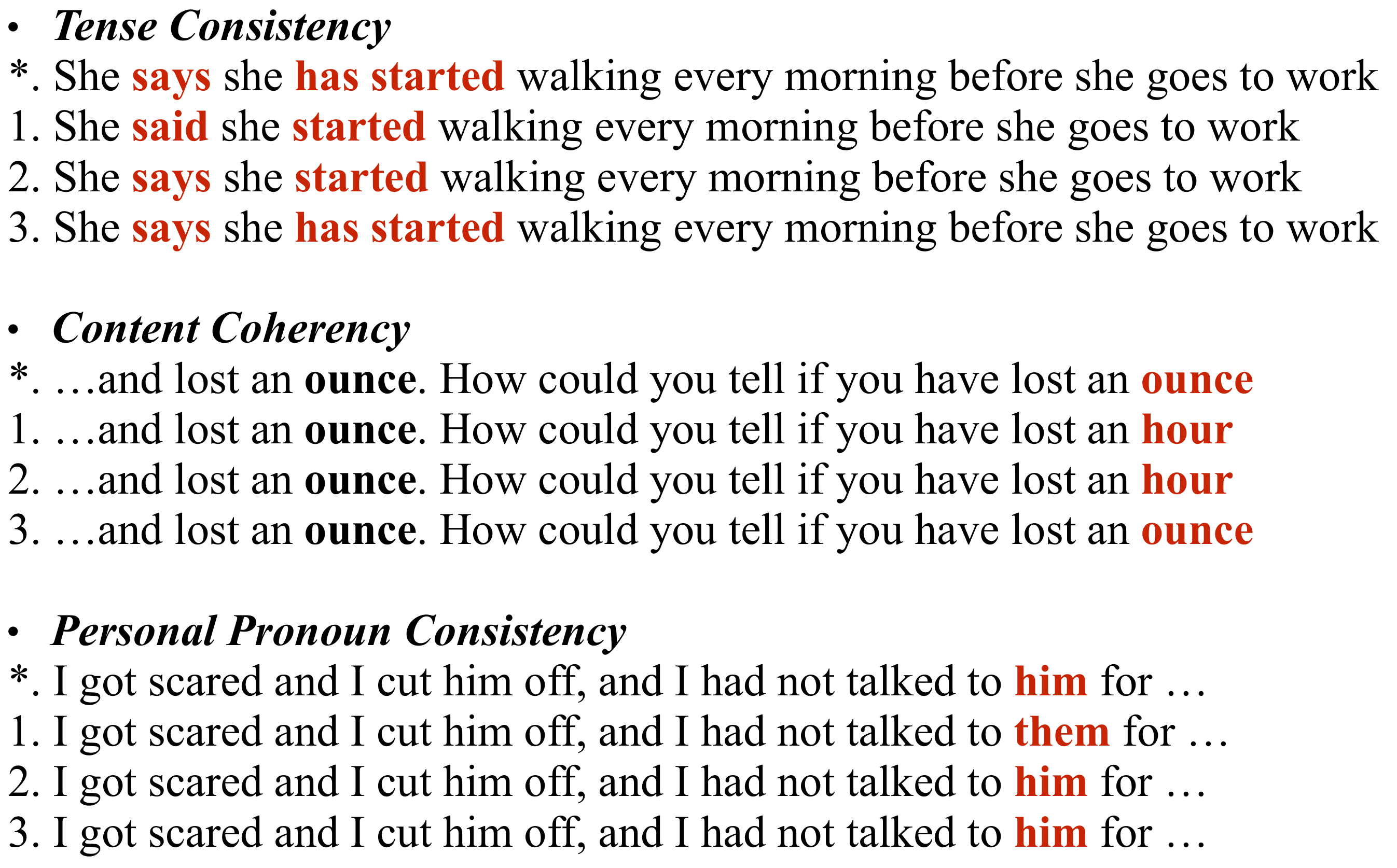}
    \caption{Examples of errors from different LMs. From top to bottom: ground truth, standard LSTM, standard LSTM with 1-best history, and self-attentive encoding CULM.}
    \label{fig:error}
    % \vspace{-0.2cm}
\end{figure}

Finally, by comparing the error pattern analysis from different systems, three examples of common improvements obtained from CULMs are summarised in Figure \ref{fig:error}.

 The most frequent error being corrected is the tense. If the surrounding context provides the correct tense, the CULM will select the utterance with matched tense in the n-best list. A second type of correction is the repetition of a word. This correction is also explicitly indicated by the high attention weight assigned to the specific word in the context when the self-attentive extraction is used. Finally, there are a couple of places where the consistency of personal pronoun use is retained.

\section{Conclusions}
\label{sec:6}

A cross-utterance LM which effectively incorporates the surrounding context information for ASR is proposed in this paper, together with two context vector extraction methods. In addition, an acoustic error sampling technique is introduced to achieve a better match between LM training and testing as well as a better generalisation to other contexts. Experiments performed on two conversational corpora showed that the proposed CULM structure outperforms the baseline in both PPL and WER, and with acoustic error sampling, further improvements were found. Specifically, the best-performing system with error sampling yielded WER reductions of 0.6\% absolute on AMI eval set, 0.3\% on the Switchboard part and 0.9\% the on Callhome part of the \emph{Eval2000} test set. Furthermore, analysis with examples of corrected error patterns demonstrated the effectiveness of the proposed method.

\newpage
\bibliographystyle{IEEEtran}

\end{document}